# Influence of Solution Efficiency and Valence of Instruction on Additive and Subtractive Solution Strategies in Humans and GPT-4

Lydia Uhler[*, a, b]

Verena Jordan[*, a]

Jürgen Buder[c]

Markus Huff[a, c]

Frank Papenmeier[a]

[a] Department of Psychology, University of Tübingen, Schleichstraße 4, 72076 Tübingen, Germany.

[b] *Current affiliation*: Department of Psychology, University of Münster, Fliednerstraße 21, 48149 Münster, Germany.

[c] Leibniz-Institut für Wissensmedien, Schleichstraße 6, 72076 Tübingen, Germany.

[*] Both authors contributed equally to this work

**Corresponding author**

Lydia Uhler, Department of Psychology, University of Münster, Fliednerstraße 21, 48149 Münster, Germany. E-Mail: lydia.uhler@uni-muenster.de

**Competing Interest:** The authors declare no competing interest.

**Funding**: This research did not receive any specific grant from funding agencies in the public, commercial, or not-for-profit sectors.

**Acknowledgements:** We thank Gerrit Anders for supporting us with collecting the GPT-4 iterations.

**Data availability:** The material, data and analysis scripts are available at https://osf.io/c78rm/.

**Preprint**: The preprint is available at https://arxiv.org/abs/2404.16692.

# Abstract

Generative artificial intelligences, particularly large language models (LLMs), play an increasingly prominent role in human decision-making contexts, necessitating transparency about their capabilities. While prior studies have shown addition biases in humans (Adams et al., 2021) and OpenAI's GPT-3 (Winter et al., 2023), this study extends the research by comparing human and GPT-4 problem-solving across both spatial and linguistic tasks, with variations in solution efficiency and valence of task instruction. Four preregistered experiments with 588 participants from the U.S. and 680 GPT-4 iterations revealed a stronger tendency towards additive transformations in GPT-4 than in humans. Human participants were less likely to use additive strategies when subtraction was relatively more efficient than when addition and subtraction were equally efficient. GPT-4 exhibited the opposite behavior, with a strong addition bias when subtraction was more efficient. In terms of valence of task instruction, GPT-4's use of additive strategies increased when instructed to "improve" (positive) rather than "edit" (neutral). These findings demonstrate that biases in human problem-solving are amplified in GPT-4, and that LLM behavior differs from human efficiency-based strategies. This highlights the limitations of LLMs and the need for caution when using them in real-world applications.

**Keywords:** Addition bias, subtraction neglect, problem-solving, generative artificial intelligence, large language model, GPT-4

## 1. Introduction

*„I have been obliged to make the present [letter] too long, for the very reason that I had not time to make it shorter."* – Blaise Pascal (1656), The Provincial Letters of Pascal

A relatively new area of research in human problem-solving concerns the use of additive and subtractive solution strategies, which was first systematically investigated by Adams et al. (2021). Problem-solving processes involve mental representations of an original state, possible transformations and different categories of actions (e.g. addition or subtraction) that can bring about these transformations. An additive solution strategy results in the transformed state having more components than the original, whereas a subtractive strategy results in the transformed state having fewer components than the original. Adams and colleagues describe that people change ideas, objects, and situations with a tendency towards additive transformations, while subtractive transformations are overlooked, even if they constitute superior solutions. This idea of *addition bias* (also referred to as *subtraction neglect*) was confirmed by Adams and colleagues in a series of studies and experiments and replicated by Fillon et al. (2022) and – to a lesser extent – by Juvrud et al. (2024).

Understanding the addition bias is crucial as it limits the consideration of alternative solutions and lead to the overuse of additive strategies. The consequences are numerous, ranging from information overload (Roetzel, 2019) to policy and regulatory overload (Hinterleitner et al., 2023), the development of slow and bloated software (Bayliss, 2022), or complex computer models with increasing numbers of parameters (Proske, 2022).

Concurrently, Large Language Models (LLMs) such as GPT-4 attracted noticeable interest due to their performance in problem-solving tasks. Nonetheless, there is a gap in the research investigating under which circumstance human problem-solving biases (e.g. *addition bias)* manifest within LLMs. This study pursued not only the investigation of human problem-solving behavior but also the question of how biases in LLMs differ from, or align with, those observed in humans. We looked explicitly at whether GPT-4 tends to favor additive strategies, in a mirror to cognitive tendencies shown by humans, or reveals patterns of behavior that are in fact unique.

**1.1 The Origins and Implications of Addition Bias**

While there is good evidence for the existence of an addition bias, its mechanisms are still being investigated. On a *cognitive* level, Adams et al. (2021) cite the overuse of heuristic memory search, in which cognitively accessible memory content is accessed quickly without sufficient consideration of alternatives. Various cognitive, cultural, and socio-ecological reasons may cause people to use addition more frequently, making it more accessible. Research showed that the likelihood of subtractive solution strategies increases when the possibility of subtractive transformations is explicitly cued in the instructions (Adams et al., 2021; Fillon et al., 2022).

Regarding the *affective* causes of addition bias, Adams and colleagues argue that additive changes are usually evaluated more positively. Terms such as "more" or "higher" are associated with "positive" or "better" (Glucksberg & McGlone, 2001). People experience more recognition for additive transformations, whereas subtractive changes are considered less creative (Meyvis & Yoon, 2021). With the notion that subtractive change is a threat to the status quo, addition bias is in line with cognitive biases such as sunk cost fallacy (Arkes & Blumer, 1985), waste aversion (Arkes, 1996), existence bias (Eidelmann et al., 2009), or loss aversion (Kahneman & Tversky, 1979). Studies on the valence of instruction (positively vs. negatively connoted task goals) in relation to addition bias yielded mixed results. Adams et al. (2021) found no differences in the frequency of subtractive strategies between the task goals of improving vs. worsening the status quo, Fillon et al. (2022) found that fewer subtractive solutions were generated in the improvement condition. In linguistic analyses, Winter et al. (2023) showed that addition-related words are used in more positive contexts than subtraction-related words. Verbs of improvement are more often associated with addition than neutral verbs of change.

Both the cognitive and affective lines of argument imply that addition bias is acquired through experience. However, Juvrud et al. (2024) demonstrated that children aged 9-10 years use subtractive transformations less frequently than adults. The origins and developmental trajectory of addition bias therefore require further research.

Since the initial investigation of addition bias, studies showed that its effects are anchored in various domains that extend far beyond the anecdotal evidence presented by Adams et al. (2021). The general disregard for subtractive solutions and the value of such manifest in a healthcare system that focuses exclusively on prescribing and developing more drugs (vs. on contemplating what could be removed or what behaviors could be stopped to make diseases disappear; Picard, 2022), in the erroneous perception of recycling (vs. reducing waste generation) as the most sustainable waste management measure (Barnett et al., 2023) or in the significantly higher likelihood of longer novels (vs. shorter novels on shortlists) to win literary awards (McGown, 2023). In view of such far-reaching effects, the question arises whether generative artificial intelligence (AI) makes more rational decisions and thus generates superior solutions to human problems.

**1.2 Large Language Models**

With the introduction of *Chat Generative Pre-Trained Transformer* (ChatGPT) in 2022, the use of and the debate about generative AI entered the public sphere. Applications range from software development and testing to writing poems, essays, business letters, or contracts (Kasneci et al., 2023). The general discussion focuses on LMMs that use deep learning for *Natural Language Processing*, (NLP) and are trained to predict the next token, i.e., individual units of text. OpenAI's GPT-4 is one of the most powerful LLMs of the current generation, although numerous other models are now available from various research organizations and companies, e.g. PaLM by Google AI, LLaMA by Meta AI or Luminous by Aleph Alpha. In contrast to its predecessors (e.g. GPT-3.5), GPT-4 is a large-scale, *multimodal* model that can process text as well as images.

**1.3 Performance and Biases of LLMs: Previous Psychological Research**

As AI systems become increasingly powerful, a lack of transparency persists regarding how these agents learn and make decisions (Gunning et al., 2019; Binz & Schulz, 2023). Researchers proposed using psychological methods to better understand the processes of the "black box" (see Ritter et al., 2017; Betz et al., 2021). Cognitive psychology, in particular, offers valuable tools for evaluating LLMs' performance and identifying biases similar to those observed in humans (see Binz & Schulz, 2023; Yax et al., 2024). Biases are implicitly present in LLMs' training data and are systematically introduced by the annotators during annotation processes (Hovy & Prabhumoye, 2021).

Several studies were conducted to systematically analyze the capabilities of LLMs using cognitive psychological methods. Early research concluded that GPT-3's performance fell behind human abilities, for example, in inductive reasoning (Han et al., 2022) and causal hypothesis formation (Kosoy et al., 2022). By contrast, Webb et al. (2023) found that the model demonstrated analogical reasoning abilities on or above human level. Binz and Schulz (2023) investigated GPT-3' performance in decision making, information search, deliberation, and causal reasoning. While no human abilities were recognizable in areas such as causal reasoning, GPT-3 achieved human-like or even better performance in other experiments. The authors point out that the well-known tasks were likely part of the training data for GPT-3. The answers could have been based on explicit knowledge of these experiments instead of actual abilities. Besides the investigation of performance, Binz and Schulz explored six cognitive biases presented by Kahneman & Tversky (1972) and found that GPT-3 mirrored three of them (framing effect, certainty effect, and overweighting bias).

With the increasing size of subsequent models, studies found improvements in performance. The analogical reasoning abilities exhibited by GPT-3 increased in GPT-4 (Webb et al., 2023). Dhingra et al. (2023) evaluated GPT-4's capabilities on a set of cognitive psychology datasets and found high levels of accuracy in different tasks. In a battery of semantic illusions and cognitive reflection tests designed to evoke biased (and incorrect) responses, both GPT-3.5 and GPT-4 outperformed humans by providing hyperrational responses, whereas earlier LLMs were unable to match human performance (Hagendorff et al., 2023). This steep shift in rationality does not hold true for other biases. Suri et al. (2024) demonstrated the prevalence of anchoring heuristics, representativeness heuristics, availability heuristics, framing effects, and endowment effects in GPT 3.5. Similarly, Yax et al. (2024) found that reasoning errors often attributed to heuristic-based human thinking also appeared in LLMs. While recent LLMs such as GPT-4 displayed fewer limitations compared to earlier versions.

**1.4 The Present Study**

The present research intended to further explore problem-solving processes, specifically the use of additive and subtractive solution strategies by humans and GPT-4, which led to three objectives:

(1) to investigate human addition bias and to replicate previous research yielding inconsistent results,

(2) to extend the research of addition bias to GPT-4, and

(3) to explore differences between humans and GPT-4 in choosing additive or subtractive strategies.

Thereby, this investigation enabled a deeper understanding of the cognitive basis of human problem-solving. If GPT-4 and humans display similar results, it may suggest that the technological foundations of LLMs closely mimic human cognition. However, if distinct differences emerge, it may highlight that human cognition involves more than just processing words, as LLMs do, suggesting a more complex, nuanced approach to problem-solving in humans.

While spatial capabilities of GPT-4 improved compared to earlier models, its performance on such tasks is still highly variable (Yamada et al., 2024). We therefore decided to use two tasks – one spatial (*symmetry task*) and one linguistic (*summary task*) – to be able to draw conclusions based on the problem type. The presence and magnitude of addition bias may be influenced by both cognitive and affective factors. To shed further light on these processes, we included two task variations. On the cognitive level, we varied *solution efficiency* in two conditions: one in which the use of additive and subtractive solution strategies was equally efficient (*addition and subtraction equally efficient* condition) and one in which the use of a subtractive strategy was more efficient than the use of an additive strategy (*subtraction more efficient* condition). On the affective level, we varied the *valence of instruction* by including one condition in which the valence of the instructive verb of the task was neutral (*valence neutral* condition) and one in which it was positive (*valence positive* condition). We conducted a series of four experiments, combining both problem types and both variations. In Experiments 1 and 2, we varied solution efficiency: in the symmetry task for Experiment 1 and in the summary task for Experiment 2. The valence of instruction was kept constant (neutral). In Experiments 3 and 4, we varied the valence of instruction while solution efficiency was kept constant (addition and subtraction equally efficient). Experiment 3 focused on the symmetry task, Experiment 4 on the summary task. Drawing on the results of previous research, we proposed three hypotheses for both humans and GPT-4:

**H1 (Addition Bias):** *Additive solution strategies are chosen significantly more often than subtractive solution strategies across all conditions.*

**H2 (Solution Efficiency):** *For both spatial and linguistic problems, the likelihood of choosing additive solution strategies is larger when the solution efficiency for addition and subtraction is equal than when the solution efficiency for subtraction is higher.*

**H3 (Valence of Instruction):** *For both spatial and linguistic problems, the likelihood of choosing additive solution strategies is larger when the valence of the instruction is positive than when the valence of the instruction is neutral.*

Regarding the third research objective, the investigation of potential differences in the problem-solving behavior of humans and GPT-4, previous research did not provide a basis for hypotheses. Studies confirmed highly accurate and rational performances of GPT-4 (see e.g. Dhingra et al., 2023; Hagendorff et al., 2023), which could lead to the assumption that addition bias is less prevalent in the LLM than in humans. Winter et al. (2023), on the other hand, showed that addition bias was not only manifested in behavioral choices but also in language itself. The analyses were extended to GPT-3 revealing that the LLM adopted more positive connotations with addition-related as compared to subtraction-related words. GPT-3 predicted the word "adding" to follow verbs of change (and even more so verbs of improvement) significantly more often than the word "removing". The results suggest a human-like addition bias to be present in GPT-4. Hence, for the present study, the exploration of differences between humans and GPT-4 in choosing additive or subtractive strategies was included as an additional, open research question:

***Research Question (RQ):*** *Do GPT-4 and humans use additive and subtractive solution strategies in the same way?*

## 2. Methodology

**2.1 Openness and Transparency**

All hypotheses, research questions, methods, analyses, and exclusion criteria of this study were preregistered. Any deviations from the preregistration are explicitly marked below. The preregistration is available on the Open Science Framework (OSF): https://osf.io/6pkwb (submitted: August 2, 2023; updated: August 9, 2023). Materials, raw data and analysis scripts are provided: https://osf.io/c78rm/.

**2.2 Participants and Data Generation**

To attain 80% power, we aimed for $n = 170$ human participants and an equivalent number of GPT-4 iterations for each of the four experiments. Human participants were recruited via Prolific, restricted to United States based individuals aged 18 or older with high English proficiency, using personal computers. GPT-4 iterations were generated via the OpenAI API with a temperature setting of 0.7 to simulate human-like answer variability. Both agents completed identical tasks under controlled conditions for direct comparison of human and LLM behaviors.

To ensure the comprehensibility of the novel linguistic symmetry task (for more details see section Materials), a test run was carried out. To address issues with externally generated responses (e.g., via ChatGPT, see Heyman & Heyman, 2023), the preregistration was updated to exclude participants whose web page lost focus during the task (via *window.blur event* method).

Data collection proceeded in two steps. As first step, data was collected until 170 participants (85 participants per condition) finished each experiment. Preliminary data filtering was carried out so that duplicates and externally assisted responses were eliminated as preregistered. Participants who did not provide informed consent or did not agree to the usage of their data were also excluded. In a second step, the retained data was analyzed for usability. Answers were excluded in case the solution entailed a final filled field count (symmetry task) or final word count (summary task) that equaled the original count of the respective problem (for details see section Procedure and Measures). After data exclusion, a second round of data collection was conducted to replace excluded participants and achieve the targeted sample sizes. The same data exclusion criteria were applied. Due to unforeseeably high exclusion rates in the symmetry task (for details see section Strengths and Limitations), targeted sample sizes were not reached after the second round of data collection. For economic reasons, no further data was collected. Participants received £0.50 for participation in the symmetry task and £1.00 in the summary task. The Ethics Committee of the Leibniz-Institut für Wissensmedien approved all procedures (LEK 2023/043).

Data from 170 GPT-4 iterations per experiment (85 iterations per condition) was collected and subsequently filtered to exclude incomplete answers. Incompletion was attributed to having reached the maximum number of tokens specified. Analogous to the human data, the remaining answers were analyzed and excluded if they could not be evaluated (i.e., unchanged filled field or word counts after transformation). In a second round of data collection, another 170 iterations per experiment were conducted. The additional answers were successively analyzed and added to the final samples until the targeted sample sizes were reached. The remaining iterations were not further considered for the study.

Sample sizes of the human and GPT-4 samples before and after data exclusion as well as the demographic characteristics of the final human samples are shown in Table 1. A detailed breakdown of data exclusions can be found in Tables A1 and A2 in the Supplementary Information (SI 1).

**Table 1**

*Summary of sample sizes and demographic characteristics*

| Experiment / Condition | Human Samples | | | | GPT-4 Samples | |
|---|---|---|---|---|---|---|
| | Sample sizes (*n*) | | Demographics [a] | | Sample sizes (*n*) | |
| | Before exclusion | Final sample | Gender (*n*) (m / f / NB)[b] | Age (years) (*M* / *SD*) | Before exclusion | Final sample |
| *Experiment 1* | | | | | | |
| Addition/subtraction equal | 191 | 67 | 33 / 32 / 2 | 33.40 / 10.89 | 107 | 85 |
| Subtraction more efficient | 124 | 65 | 30 / 31 / 4 | 34.54 / 10.97 | 94 | 85 |
| Total | 315 | 132 | 63 / 63 / 6 | 33.96 / 10.90 | 201 | 170 |
| *Experiment 2* | | | | | | |
| Addition/subtraction equal | 107 | 87 | 39 / 45 / 2 | 36.90 / 12.30 | 91 | 85 |
| Subtraction more efficient | 99 | 84 | 40 / 43 / 1 | 37.58 / 12.41 | 86 | 85 |
| Total | 206 | 171 | 79 / 88 / 3 | 37.23 / 12.32 | 177 | 170 |
| *Experiment 3* | | | | | | |
| Valence neutral | 178 | 60 | 31 / 27 / 2 | 35.48 / 10.67 | 111 | 85 |
| Valence positive | 174 | 59 | 34 / 23 / 2 | 34.69 / 10.96 | 111 | 85 |
| Total | 352 | 119 | 65 / 50 / 4 | 35.09 / 10.77 | 222 | 170 |
| *Experiment 4* | | | | | | |
| Valence neutral | 107 | 83 | 43 / 37 / 2 | 37.59 / 11.60 | 89 | 85 |
| Valence positive | 121 | 83 | 36 / 46 / 1 | 36.43 / 12.39 | 86 | 85 |
| Total | 228 | 166 | 79 / 83 / 3 | 37.01 / 11.98 | 175 | 170 |

[a] Final samples only
[b] Deviations from final sample sizes are due to some participants' nondisclosure of gender

## 2.3 Materials

Both tasks of the study are adapted versions of Adams et al. (2021). The *symmetry task* is a linguistic version of their grid task representing the *spatial* problem of this study. Participants were presented a grid with a non-symmetrical field pattern. A field could either be filled ("[X]") or empty ("[ ]"). They were then asked to describe how to change the pattern to make it "perfectly symmetrical from left to right and from bottom to top" by switching fields from "[X]" to "[ ]" or from "[ ]" to "[X]" with as little switching as possible. A smaller grid (4-by-4) was used to account for the increased complexity of solving this task by articulating the proposed solution instead of interactively clicking on the grid. The test run of the task revealed that the original verb "toggle" was not comprehensible for most participants which is why "switch" was used instead.

In Experiment 1, the *solution efficiency* variation was implemented by either presenting an initial field pattern for which the most efficient additive and subtractive solutions required the same amount of switching (addition and subtraction equally efficient condition) or an initial field pattern for which the most efficient solution was subtraction (subtraction more efficient condition). In Experiment 3, the *valence of instruction* was varied by exchanging the instructive verb of the task from "change" (valence neutral condition) to "improve" (valence positive condition). Figure 1 illustrates the different variations used in the symmetry task.

**Figure 1**

*Overview of variations used in the symmetry task in Experiments 1 and 3*

| Variation: | Exp. 1: Solution efficiency | | Exp. 3: Valence of instruction | |
|---|---|---|---|---|
| **Condition:** | addition and subtraction equally efficient | subtraction more efficient | valence neutral | valence positive |
| **Initial field pattern:** | A B C D<br>1 [X][ ][ ][ ]<br>2 [ ][X][X][ ]<br>3 [ ][X][X][ ]<br>4 [X][ ][ ][ ] | A B C D<br>1 [X][ ][ ][ ]<br>2 [ ][X][X][ ]<br>3 [ ][X][X][ ]<br>4 [ ][ ][ ][ ] | A B C D<br>1 [X][ ][ ][ ]<br>2 [ ][X][X][ ]<br>3 [ ][X][X][ ]<br>4 [X][ ][ ][ ] | A B C D<br>1 [X][ ][ ][ ]<br>2 [ ][X][X][ ]<br>3 [ ][X][X][ ]<br>4 [X][ ][ ][ ] |
| **Instruction:** | "Your task is to *change* the pattern…" | "Your task is to *change* the pattern…" | "Your task is to *change* the pattern…" | "Your task is to *improve* the pattern…" |

*Note.* Red circle as well as red and/or bold fonts serve illustrative purposes only and were not present in the actual experiments.

The *summary task* constitutes the *linguistic* problem of the study. Participants were provided a newspaper article as well as a corresponding summary and were then asked to edit the summary. While the task itself was very similar to the one presented by Adams et al. (2021), a different, publicly available text (see Dole, 2015) and summary (see Zentrum für Schulqualität und Lehrerbildung, 2016) were used. In Experiment 2, *solution efficiency* was varied by providing the word count of the initial summary and by specifying word count ranges that were prohibited for the edited summaries. While the prohibited range in the addition and subtraction equally efficient condition was +/- 12 words around the original word count, the subtraction more efficient condition was operationalized by shifting the range so that getting outside the range required subtracting at least six words or adding at least 18 words. In Experiment 4, the variation of valence of instruction was implemented by exchanging the instructive verb "edit" used in the neutral condition by the verb "improve" in the positive condition. An overview of the variations used in the summary task is provided in Figure 2. The materials of all experiments can be found in the Supplementary Information (SI 2).

**Figure 2**

*Overview of variations used in the summary task in Experiments 2 and 4*

| Variation: | Exp. 2: Solution efficiency | | Exp. 4: Valence of instruction | |
|---|---|---|---|---|
| **Condition:** | addition and subtraction equally efficient | subtraction more efficient | valence neutral | valence positive |
| **Instruction:** | "We would like you to **edit** this summary." | "We would like you to **edit** this summary." | "We would like you to **edit** this summary." | "We would like you to **improve** this summary." |
| **Prohibited word count range:** | "The summary currently contains 76 words. The **edited** summary MUST NOT be **between 64 and 88 words** long." | "The summary currently contains 76 words. The **edited** summary MUST NOT be **between 70 and 94 words** long." | "The summary currently contains 76 words. The **edited** summary MUST NOT be **between 64 and 88 words** long." | "The summary currently contains 76 words. The **improved** summary MUST NOT be **between 64 and 88 words** long." |

*Note.* Red and/or bold fonts serve illustrative purposes only and were not present in the actual experiments.

**2.4 Procedure and Measures**

For the human sample, a survey was constructed containing questions regarding the participants' gender and age as well as the experiment-specific task in one of the two conditions (between-subject design). After indicating their consent to the study prerequisites, participants were randomly assigned to one condition. Each participant was allowed to participate in only one of the four experiments. Experiments were conducted sequentially, allowing exclusion of previous participants for subsequent experiments. Before completing the assigned task, participants were instructed neither to leave the website nor to use external help to solve the problem. After survey completion, participants received a debriefing on the study purposes and were then asked to give their final consent to data usage. GPT-4 iterations were generated using the OpenAI API, the temperature parameter at 0.7 and the maximum number of tokens set to 500. GPT-4 was instructed to be "a helpful assistant" (default).

After data collection, the data of Experiments 2 and 4 (summary task) was analyzed and coded via R. For Experiments 1 and 3 (symmetry task), an automated evaluation of the answers was not possible. Analysis and coding were performed manually by one of the study authors. The retained solutions in the final samples were categorized as either additive or subtractive solution strategies regardless of the accuracy of the answer given regarding the respective task. For the symmetry task, a solution strategy was considered additive (vs. subtractive) if the final filled field count after the proposed transformation was higher (vs. lower) than the original filled field count. Correspondingly, in the summary task, a higher (vs. lower) final word count than the original was regarded as additive (vs. subtractive). As preregistered, the accuracy of the solutions was evaluated separately for additional analyses on correct solutions only. Answers were considered as accurate if the proposed solution resulted in a grid pattern which was symmetrical from left to right and from top to bottom (symmetry task) or if the word count of the edited text was outside the specified prohibited ranges (summary task).

### 2.5 Analytic Strategy

All data analyses were performed using R. All tests were two-tailed with an alpha level of .05. Testing the addition bias across all conditions (H1), a chi-square test was performed to determine whether the total frequency of additive solution strategies significantly differed from the total frequency of subtractive strategies for all solutions regardless of their accuracy. A chi-square test of homogeneity for the 2 (agent: human vs. GPT-4) x 2 (frequency of solution strategy: additive vs. subtractive) contingency table and the corresponding Phi coefficient were calculated to investigate whether humans and GPT-4 use additive and subtractive strategies in the same manner (RQ). Logistic regression analyses were conducted separately for each experiment to explain the choice of solution strategies (additive vs. subtractive). Type-II ANOVAs were calculated using the *Anova()* function in the *car* package (Fox & Weisberg, 2019). Analyses were performed on all retained answers of the final samples including those which presented incorrect solutions to the respective problem. Chi-square tests and logistic regression analyses were chosen due to categorical dependent variables (frequency of solution strategy: additive vs. subtractive).

To test whether the likelihood of choosing additive solution strategies was larger when the solution efficiency for addition and subtraction was equal as compared to the solution efficiency being higher for subtraction (H2), binomial generalized linear models with logit link function were calculated for Experiments 1 and 2. These models included the main effects of both agent (human vs. GPT-4) and solution efficiency (addition and subtraction equally efficient vs. subtraction more efficient) as well as their interaction as predictors. In the same manner, binomial generalized linear models including the main effects of agent (human vs. GPT-4) and valence of instruction (neutral vs. positive) as well as their interaction were computed for Experiments 3 and 4 to analyze whether the likelihood of choosing additive solution strategies was larger when the valence of the instruction was positive as compared to neutral (H3). In case of significant interaction effects, post-hoc simple main effects were calculated for humans and GPT-4 separately to explore the agents' respective patterns of solution strategies chosen in the different conditions (RQ).

To evaluate whether solution accuracy makes a difference in the types of solution strategies chosen (i.e., well thought out solutions being less prone to biases (see Kahneman, 2011)), all analyses were preregistered to also be conducted on the subsamples of solely accurate solutions.

## 3. Results

### 3.1 Hypothesis 1: Cross-Sample Presence of Addition Bias

Across all 1268 trials included in the final samples of the four experiments, the number of additive solution strategies proposed was 816 (64.35%). Subtractive solution strategies were chosen in 452 trials (35.65%). The total frequency of additive solution strategies significantly differed from the total frequency of subtractive solution strategies, $X^2(1, N = 1,268) = 104.49, p < .001$.

The explorative calculation of the chi-square test of homogeneity for the 2 (agent) x 2 (frequency of solution strategy) contingency table as well as the corresponding Phi coefficient showed a significant association between agent (human vs. GPT-4) and the frequencies of chosen solution strategies, $X^2(1, N = 1,268) = 26.04, p < .001, \varphi = 0.14$, with humans choosing additive solution strategies relatively less often than GPT-4 (56.97% vs. 70.74%).

**3.2 Hypothesis 2: Impact of Solution Efficiency on Choice of Additive Solution Strategy**

In Experiment 1 (symmetry task), the logistic regression analysis showed a significant interaction effect between agent (human vs. GPT-4) and solution efficiency, $X^2(1, n = 302) = 20.54, p < .001$, a significant main effect of agent, $X^2(1, n = 302) = 6.02, p = .014$, but a non-significant main effect of condition, $X^2(1, n = 302) = 1.67, p = .196$. The post-hoc analysis of the simple main effects revealed that the association between the solution efficiency for the respective task and the choice of additive solution strategies was significant but contrary for humans and GPT-4. While humans expectedly chose additive solution strategies significantly more often in the addition and subtraction equally efficient condition as compared to the subtraction more efficient condition (65.67% vs. 29.23%), $X^2(1, n = 132) = 19.95, p < .001$, the pattern of relative frequencies was reversed for GPT-4 (54.12% vs. 69.41%), $X^2(1, n = 170) = 4.27, p = .039$. The results of Experiment 1 are depicted in Figure 3. The beta coefficients and corresponding odds ratios of the logistic regression analysis are summarized in Table C1 in the Supplementary Information (SI 3).

The corresponding analyses of the summary task data in Experiment 2 indicated similar results with a significant interaction effect between agent (human vs. GPT-4) and solution efficiency, $X^2(1, n = 341) = 10.84, p < .001$, a significant main effect of agent, $X^2(1, n = 341) = 50.30, p < .001$, and a non-significant main effect of condition, $X^2(1, n = 341) = 0.32, p = .574$. Exploring the simple main effects, humans and GPT-4 again showed significant but contradicting results regarding the association between solution efficiency and the type of solution strategies chosen. The relative frequency of additive solution strategies chosen by humans was significantly lower in the subtraction more efficient condition (57.47% vs. 39.29%), $X^2(1, n = 171) = 5.78, p = .016$, whereas the respective relative frequency was significantly higher for GPT-4 (77.65% vs. 90.59%), $X^2(1, n = 170) = 5.44, p = .020$. Figure 3 also shows the interaction plot of Experiment 2. Beta coefficients and odds ratios can be found in Table C2 in the Supplementary Information (SI 3).

**Figure 3**

*Percentage of chosen additive solution strategy depending on solution efficiency: GPT-4 vs. humans in symmetry task (Experiment 1) and in summary task (Experiment 2).*

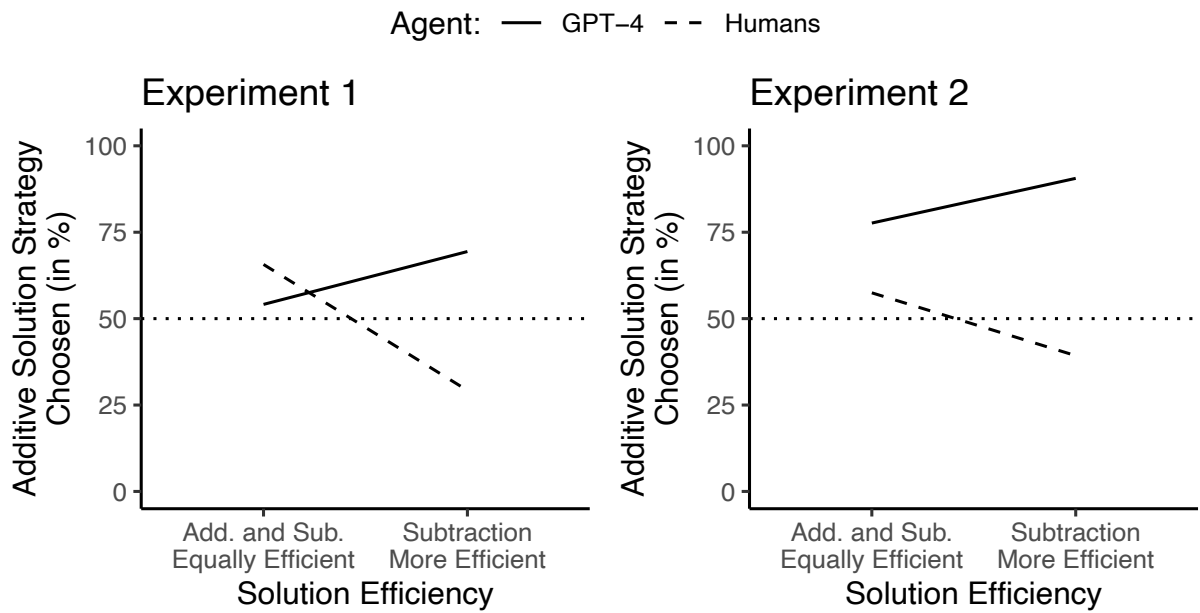

condition was therefore neither significant for humans (65.55% vs. 61.02%, *n* = 119) nor for GPT-4 (45.88% vs. 55.29%, *n* = 170). Across both conditions, humans chose additive strategies significantly more often than GPT-4. The results of Experiment 3 are illustrated in Figure 4. The summary of beta coefficients and odds ratios can be found in Table C3 in the Supplementary Information (SI 3).

While the variation of solution efficiency showed similar effects in the symmetry and the summary task, the results of the summary task differed from the symmetry task when the valence of instruction was varied. In Experiment 4, the logistic regression analysis showed a significant interaction effect between agent (human vs. GPT-4) and valence of instruction, $X^2(1, n = 336) = 6.84, p = .009$, as well as significant main effects of both agents, $X^2(1, n = 336) = 15.20, p < .001$, and condition, $X^2(1, n = 336) = 10.59, p = .001$. The calculation of the simple main effects revealed that the association between valence of instruction and the relative frequency of choosing additive solution strategies was only significant for GPT-4 (76.47% vs. 96.47%), $X^2(1, n = 170) = 15.70, p < .001$, but not for humans (65.06% vs. 73.49%), $X^2(1, n = 166) = 1.38, p = .240$. Figure 4 also shows the results of Experiment 4. For beta coefficients and odds ratios, see Table C4 in the Supplementary Information (SI 3).

**Figure 4**

*Percentage of chosen additive solution strategy depending on valence of instruction: GPT-4 vs. humans in symmetry task (Experiment 3) and in summary task (Experiment 4)*

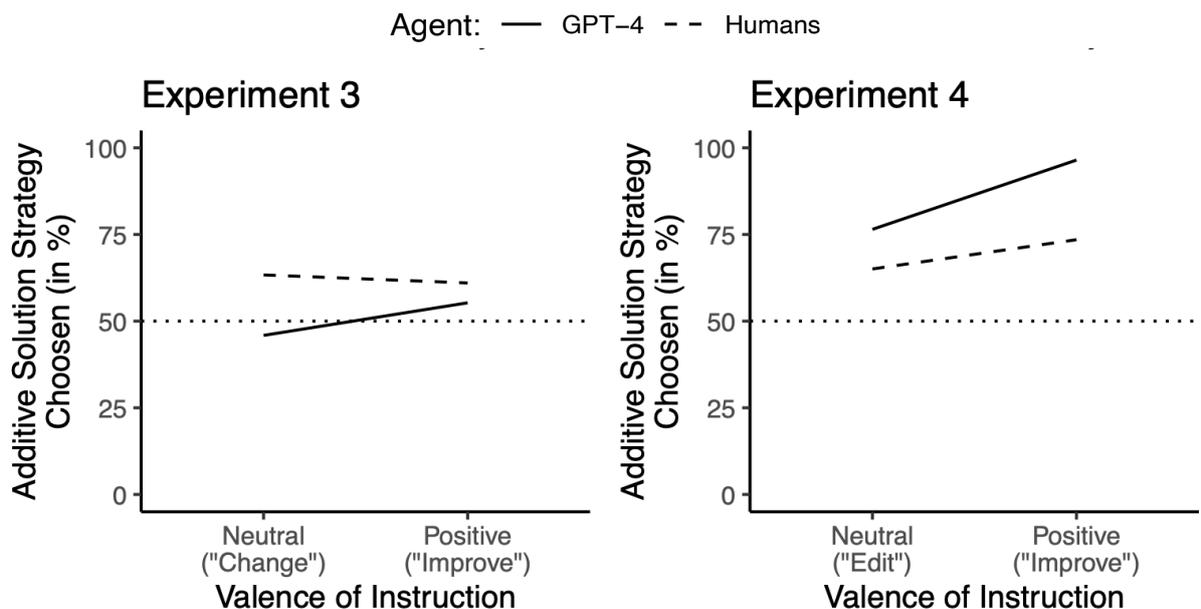

### 3.4 Additional Analyses

3.4.1 Analyses of Accurate Solutions

As preregistered, all analyses were additionally planned to be conducted on the subsamples of solely accurate solutions. However, the evaluation of solution accuracy revealed that the distribution of accurate and inaccurate solutions differed tremendously between humans and GPT-4 (see Table 2). While the share of correct solutions proposed by humans was at least 80% in each experiment, there were hardly any accurate solutions on the part of GPT-4 in the symmetry task and mainly shares below 20% in the summary task. Separate analyses for accurate solutions only were therefore not feasible.

**Table 2**

*Overview of accuracy of solutions proposed by humans and by GPT-4*

| Experiment / Condition | Human Samples | | GPT-4 Samples | |
|---|---|---|---|---|
| | Accurate Solutions (*n*) | Inaccurate Solutions (*n*) | Accurate Solutions (*n*) | Inaccurate Solutions (*n*) |
| *Experiment 1* | | | | |
| Addition/subtraction equal | 65 (97.01%) | 2 (2.99%) | 2 (2.35%) | 83 (97.65%) |
| Subtraction more efficient | 52 (80.00%) | 13 (20.00%) | 2 (2.35%) | 83 (97.65%) |
| *Experiment 2* | | | | |
| Addition/subtraction equal | 74 (85.06%) | 13 (14.94%) | 15 (17.65%) | 70 (82.35%) |
| Subtraction more efficient | 73 (86.90%) | 11 (13.10%) | 13 (15.29%) | 72 (84.71%) |
| *Experiment 3* | | | | |
| Valence neutral | 58 (96.67%) | 2 (3.33%) | 1 (1.18%) | 84 (98.82%) |
| Valence positive | 55 (93.22%) | 4 (6.78%) | 3 (3.53%) | 82 (96.47%) |
| *Experiment 4* | | | | |
| Valence neutral | 68 (81.93%) | 15 (18.07%) | 15 (17.65%) | 70 (82.35%) |
| Valence positive | 72 (86.75%) | 11 (13.25%) | 57 (67.06%) | 28 (32.94%) |

*Note.* Solutions were considered as accurate if the proposed solution in the symmetry task resulted in a grid pattern which was perfectly symmetrical from left to right and from bottom to top (Experiments 1 and 3) or if the word count of the edited text in the summary task was outside the prohibited ranges as specified (Experiments 2 and 4).

## 4. Discussion

Humans prefer additive transformations when changing ideas, objects, and situations, while subtractive transformations are overlooked even if they represent superior solutions. This addition bias was only recently systematically investigated (see Adams et al., 2021) but – as the opening quote may imply – probably existed for centuries with implications on today's everyday life (see e.g., Barnett et al., 2023; McGown, 2023; Picard, 2022).

This study followed three objectives by adopting three hypotheses and an additional research question. The results confirmed a general prevalence of addition bias across all conditions and both, humans and GPT-4 (H1). However, further analysis of solution strategies chosen by humans vs. GPT-4 showed that the patterns in strategy choices differed significantly between the agents (RQ). GPT-4 chose additive strategies more frequently than humans (70.74% vs. 56.97%). A higher likelihood of choosing additive strategies when additive and subtractive solutions are equally efficient (vs. when subtractive solutions are more efficient) could not be found (H2). Instead, the results in the spatial (Experiment 1) and the linguistic task (Experiment 2) indicated that the variation of solution efficiency had contrary effects on humans compared to GPT-4 (RQ). While human's choice of solution strategies supported the hypothesis, GPT-4 showed an increased likelihood of additive solutions when the solution efficiency was higher for subtractive solutions. Similarly, no support for the general assumption could be found that the choice of additive strategies is more likely when the valence of task instruction is positive as compared to neutral (H3). In the spatial task (Experiment 3), the variation of the valence had no effect on either humans or GPT-4 (RQ). In the linguistic task (Experiment 4), on the other hand, varying the valence of instruction worked differently on both agents (RQ). While GPT-4's responses supported the hypothesis by more frequently choosing additive strategies when positively instructed, again, no effect of the valence of instruction was found in humans.

The first research objective was to further investigate human addition bias and to replicate previous research. The results partially confirmed the findings of Adams et al. (2021). When it was more efficient to choose subtractive solution strategies, no addition bias could be found in humans, like in Adams and colleagues' grid task. However, due to the adaption of the symmetry task, a linguistic cue for subtraction ("switch any field from '[X]' to '[ ]' or from '[ ]' to '[X]'") was given, which was not the case in the original task. Similarly, operationalizing the solution efficiency variation in the summary task by specifying prohibited word count ranges also provided an originally non-existent cue for subtractive solutions. Given the significant impact of subtraction cues found by Adams et al. (2021) and Fillon et al. (2022), these task deviations might explain the differences in the results. Despite the subtraction cues, addition bias was found in humans when the solution efficiency was equal for additive and subtractive solutions. As the study did not include any no-cue control groups, the effect of the cues provided – especially in conjunction with the variation of solution efficiency – cannot be further explored. In line with Adams et al. (2021) but in contrast to Fillon et al. (2022) and the linguistic findings of Winter et al. (2023), there was no effect of the valence of task instruction regarding human solution strategy choices in either the spatial or the linguistic task. This may imply that despite the existence of differences between verbs of change and verbs of improvement in their relation to addition in the English language (Winter et al., 2023), such a subtle task variation does not suffice to trigger different behavioral outcomes in humans.

Secondly, the present study aimed to extend addition bias research to GPT-4. The results indicated that GPT-4 displayed a higher likelihood for additive solution strategies across almost all conditions in both the spatial and the linguistic task. This preference was especially pronounced in the linguistic task, which is better suited to the capabilities of the LLM. These results confirmed the linguistic findings of Winter et al. (2023) for GPT-3. Therefore, it can be concluded that addition bias falls in line with other cognitive biases found in LLMs.

The third research objective was to explore differences between humans and GPT-4 in choosing additive or subtractive solution strategies. Results revealed significant differences between the agents in unexpected manners. Across all trials and in three out of four experiments, GPT-4 displayed an even higher preference for additive solution strategies than humans. This indicated that the addition bias implicitly present in human language (see Winter et al., 2023) is even reinforced by GPT-4 (see Ntoutsi et al., 2020). In contrast to humans, GPT-4 acted irrationally when solution efficiency was varied in that additive solutions were chosen even more frequently when the solution efficiency suggested subtractive solutions. This behavioral pattern was surprising. For the symmetry task, a potential explanation could be the influence of an inherent anchoring bias (see Suri et al., 2024) leading to a higher likelihood of adding filled fields when the baseline number of filled fields is smaller (like in the subtraction more efficient condition as compared to the addition and subtraction equally efficient condition; Fischer et al., 2021). For the linguistic task, the valence of instruction influenced GPT-4's solution strategy choices which was not the case for humans. This might be in line with the argument that human linguistic associations – in this case the stronger relation between addition and verbs of improvement as compared to verbs of change (Winter et al., 2023) – are not only mirrored but even amplified by LLMs (Ntoutsi et al., 2020).

**4.1 Strengths and Limitations of the Study**

This study exhibits several strengths in terms of both content and methodology. The subject of additive and subtractive solution strategies is highly relevant as it was investigated systematically only in the past few years with mixed results. While the original study by Adams et al. (2021) may have been part of the LLM's training data, knowledge about the materials and confounded solution suggestions can be ruled out in this study as the tasks were significantly adapted by the authors. By presenting the exact same tasks to both humans and the LLM, influencing factors other than language and the linguistic variations used were minimized.

Important methodological limitations of this study must be considered. While presenting the agents the exact same linguistically articulated tasks is important to draw comparative conclusions, humans may not be used to non-interactive problem types, like the symmetry task. Many participants merely repeated the task instruction in their own words or theoretically described how they would proceed without describing an actual solution. This was not the case when interactive versions of the task were used in previous studies (see Adams et al., 2021; Juvrud et al., 2024). Difficulty in task comprehension may have led to a higher cognitive load which, in itself, has been shown to influence addition bias (Adams et al., 2021). During data analysis, it became clear that despite trying to reduce linguistic ambiguity, many participants understood "switch fields around" instead "switch fields on and off", resulting in unchanged filled field counts and, hence, further data exclusions. GPT-4 had less issues understanding *how* the task should be performed but failed to understand *what* the task asked to produce. The LLM seemed to lack any comprehension of symmetry, as reflected in the low rates of accurate solutions, confirming earlier research on the spatial abilities of LLMs (see e.g., Yamada et al., 2024). Additionally, GPT-4 mostly failed to produce correct solutions in the summary task. One explanation might be the difficulty of LLMs to understand negations (Garcia-Ferrero et al., 2023), which was necessary to accurately solve the task ("The edited summary MUST NOT be between 64 and 88 words long."). These low problem-solving abilities stand in contrast to earlier research results (see e.g., Dhingra et al., 2023; Hagendorff et al., 2023; Webb et al., 2023). It is debatable whether the analysis of a problem-solving bias is meaningful considering such an inability to solve tasks as specified.

Fischer et al. (2021) criticized the tasks and execution by Adams et al. (2021), several aspects hold true for this study. First, the order of cues for addition and subtraction was not counterbalanced. While, in contrast to Adams and colleagues' cues and to general convention, the implicit cue for subtraction was given first in both tasks making subtraction more salient, no further conclusions can be drawn regarding a possible sequence effect. A second point of criticism, as already pointed out, is that the baseline numbers of filled and empty fields in the symmetry task were not equal. This design biases against subtraction and makes addition more likely. Third, the patterns of filled fields presented in the symmetry task were not controlled for widespread cognitive spatial associations (addition is associated with right/up, subtraction with left/down; Winter et al., 2015). This might have increased the likelihood of additive solutions. Finally, the tasks used were similarly complex as the original tasks. According to Fischer et al. (2021), this task complexity mixes processing domains which can lead to confounding by cross-domain associations.

While various more or less known cognitive and emotional processes lead to human behavior, GPT-4 shows behavior in form of verbal responses with unknown underlying processes. In their commentary on psychological research on AI, Shiffrin and Mitchell (2023) questioned which aspects of LLM responses should be investigated in comparative studies with humans – the verbal responses, the token probabilities, or the neural networks' internal representations. Without further consideration, however, this study followed previous cognitive psychological research and opted for verbal responses.

**4.2 Constraints on Generality**

The present study entails constraints on generality regarding both humans and GPT-4. Participants were only recruited from the United States which limits the applicability of the results to other cultural contexts, as previous research showed that addition bias varied across cultures (Juvrud et al., 2024). Due to LLM's context dependence implying that small linguistic changes in prompts could lead to entirely different outcomes (Binz & Schulz, 2023; Shiffrin & Mitchell, 2023) generalizing the results is challenging. Instead of the deterministic approach, temperature parameter at 0, chosen by most GPT studies, the parameter was kept at 0.7 to attain variability in the answers. This way, a "sample" of GPT-4 trials could be collected. However, the parameter represents a black box and the assumption that the variability at 0.7 resembles human variability is highly contestable. It remains unclear how to correctly interpret the results and whether they tell us anything about the "inner processes" of GPT-4 at all. Finally, as Shiffrin & Mitchell (2023) remark, "psychological assessments designed to test humans' higher-level cognitive abilities […] may in fact not test these abilities at all in LLMs, even when LLMs – trained on huge swaths of human-generated text – produce similar responses as humans" (p. 2). In other words, the interpretation of results obtained from LLMs in human terms is questionable and, in the long-term, may even become less and less feasible as AI models emerge (Schwartz, 2022). Any generalizations must be made with caution.

### 4.3 Considerations for Future Research

The findings revealed human addition bias in certain settings and a more pronounced addition bias in GPT-4 as compared to humans. Future research could incorporate less complex tasks as proposed by Fischer et al. (2021) to explore differences between the agents on a more granular level. To obtain a better understanding of the interplay of solution efficiency and subtraction cues, control groups should be included. To gain deeper insights into the inner processes of humans and LLMs, more profound cognitive psychological methods may be useful, like the Cognitive Reflection Test (Frederick, 2005). For LLMs, future research may apply chain-of-thought-prompting, which has been shown to improve LLM reasoning abilities (see e.g., Betz et al., 2021; Wei et al., 2022) and could make the problem-solving process more transparent and thereby comparable. For humans, think-aloud protocols (Lewis, 1982) provide a better sense of the task comprehension, solution generation and solution choice processes and, hence, a better idea of where along the way addition bias occurs. The use of such methods also shed light on which factors other than language influence human behavior and, hence, differentiate humans from LLMs. Within the NegativePrompt framework, Wang et al. (2024) demonstrated that negative emotional stimuli can positively influence LLM's performance. This highlights the importance of understanding not only cognitive but also emotional factors in interaction of humans and LLMs. Integrating frameworks like NegativePrompt (Wang et al., 2024) into the investigation might not only enrich our understanding of LLM biases but emphasize the potential for emotional engagement strategies to enhance LLM performance across various applications. Finally, other cognitive biases, such as the sunk cost fallacy or waste aversion, were mentioned as possible underlying factors of addition bias (see Meyvis & Yoon, 2021). This study has not provided further insights if and how the addition bias differs from these concepts. Therefore, future research could control for other cognitive biases to delineate addition bias more clearly.

**4.4 Conclusion**

To conclude, this study provides further evidence for a human preference for additive solutions under certain circumstances and initial evidence for an even more pronounced addition bias in GPT-4. While humans did not exhibit addition bias when subtractive solutions were superior in terms of efficiency, GPT-4 showed a higher likelihood of choosing additive solution strategies in this scenario as compared to additive and subtractive solutions being equally efficient. This irrational behavior needs further investigation. Additionally, the valence of the task instruction did not affect human addition bias, whereas GPT-4 displayed an even stronger preference for additive solutions in the linguistic task when asked to improve compared to edit. These findings indicate that addition bias has been linguistically transferred to the LLM that has not only adopted but even amplified this preference for addition by magnifying subtle relations in the English language. This is important to keep in mind when seeking the assistance of AI in solving real-world problems.